\documentclass[10pt,twocolumn,letterpaper]{article}

\usepackage[pagenumbers]{iccv} 

\usepackage{algorithm}
\usepackage{algorithmic}

\definecolor{iccvblue}{rgb}{0.21,0.49,0.74}
\usepackage[pagebackref,breaklinks,colorlinks,allcolors=iccvblue]{hyperref}
\usepackage{multirow}
\usepackage{makecell}

\def\paperID{10878} 
\def\confName{ICCV}
\def\confYear{2025}

\title{RoBridge: A Hierarchical Architecture Bridging Cognition and Execution for General Robotic Manipulation}

\author{
Kaidong Zhang\textsuperscript{1} \quad
Rongtao Xu\textsuperscript{2,5} \quad
Pengzhen Ren\textsuperscript{3} \quad
Junfan Lin\textsuperscript{3} \quad
\\
{
Hefeng Wu\textsuperscript{1}\quad
Liang Lin\textsuperscript{1,3,4}\footnotemark[2]\quad
Xiaodan Liang\textsuperscript{1,2,3}\footnotemark[2]}
\\
\resizebox{0.7\linewidth}{!}{
\textsuperscript{1}Sun Yat-sen University \quad 
\textsuperscript{2}MBZUAI \quad 
\textsuperscript{3}Peng Cheng Laboratory \quad 
\textsuperscript{4}X-Era.AI Inc. \quad
\textsuperscript{5}Spatialtemporal AI
}\\
\color{red}{https://abliao.github.io/RoBridge/}
}

\begin{document}
\maketitle
\renewcommand{\thefootnote}{\fnsymbol{footnote}} 
\footnotetext[2]{Corresponding authors} 

\begin{abstract}

Operating robots in open-ended scenarios with diverse tasks is a crucial research and application direction in robotics. While recent progress in natural language processing and large multimodal models has enhanced robots' ability to understand complex instructions, robot manipulation still faces the procedural skill dilemma and the declarative skill dilemma in open environments. Existing methods often compromise cognitive and executive capabilities.  To address these challenges, in this paper, we propose RoBridge, a hierarchical intelligent architecture for general robotic manipulation. It consists of a high-level cognitive planner (HCP) based on a large-scale pre-trained vision-language model (VLM), an invariant operable representation (IOR) serving as a symbolic bridge, and a guided embodied agent (GEA). RoBridge maintains the declarative skill of VLM and unleashes the procedural skill of reinforcement learning, effectively bridging the gap between cognition and execution. RoBridge demonstrates significant performance improvements over existing baselines, achieving a 75\% success rate on new tasks and an 83\% average success rate in sim-to-real generalization using only five real-world data samples per task. This work represents a significant step towards integrating cognitive reasoning with physical execution in robotic systems, offering a new paradigm for general robotic manipulation.

\end{abstract}    
\section{Introduction}
\label{sec:intro}

{

}


\begin{figure*}
    \centering
    \includegraphics[width=\linewidth]{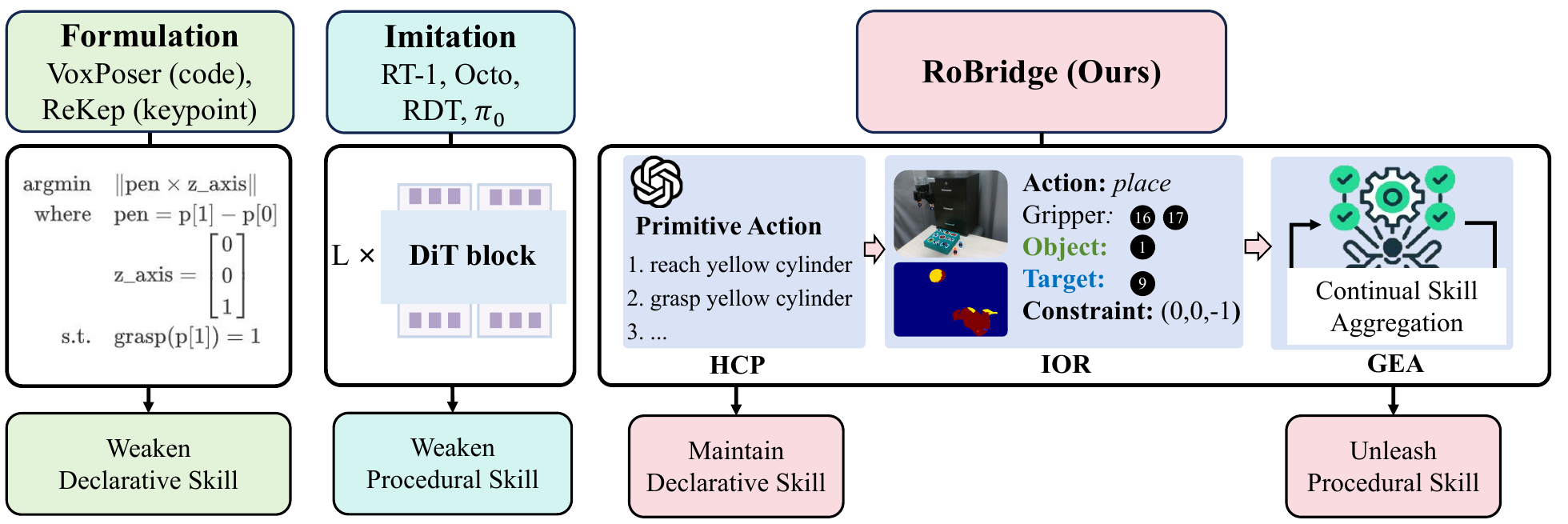}
    \vspace{-1.2em}
    \caption{\textbf{Comparison of RoBridge and previous methods}. Declarative skill methods (left) directly generate specific control commands in a formulaic way, such as determining trajectories by minimizing cost. However, due to a lack of interaction experience with the physical world, the generated commands are often incorrect. Procedural skill methods (middle) forcibly transform a vision-language model (VLM) into a robotics model using a data-driven approach, but it is not effective in dealing with unseen situations. Our method, RoBridge(right), enables the VLM to generate physically intuitive representations as a symbolic bridge. This symbolic bridge is characterized by its invariance, allowing it to communicate with the underlying embodied agent in a universal manner. Meanwhile, the embodied agent continuously interacts with the physical world to gain continual skill aggregation, fully leveraging the strengths of both the VLM and reinforcement learning.
    }
    \label{fig:comparison}
    \vspace{-0.4em}
\end{figure*}

Operating robots in open-ended scenarios with diverse tasks is a significant research and application direction in the field of robotics. In such scenarios, robots are required to understand natural language instructions and accurately execute complex tasks while adapting to dynamic environmental changes and uncertainties \cite{hu2024generalpurposerobotsfoundationmodels,duan2022surveyembodiedaisimulators}. In recent years, the rapid development of natural language processing and large multimodal models \cite{openai2024gpt4technicalreport, touvron2023llamaopenefficientfoundation, alayrac2022flamingovisuallanguagemodel}  has led to substantial progress in robots' ability to understand and execute complex instructions. However, robot manipulation still faces two major challenges when dealing with complex instructions in open environments: the procedural skill dilemma and the declarative skill dilemma.
\begin{itemize}
    \item \textbf{Procedural Skill Dilemma.} To acquire the ability to manipulate objects according to instructions, embodied models\cite{rt12022arxiv,octo_2023,liu2024rdt1bdiffusionfoundationmodel,black2024pi0visionlanguageactionflowmodel} like RDT \cite{liu2024rdt1bdiffusionfoundationmodel} and $\pi_0$ \cite{black2024pi0visionlanguageactionflowmodel} typically employ data-driven trajectory fitting methods. However, these methods frequently suffer catastrophic performance degradation when confronted with environmental variations, including fluctuating illumination conditions, camera pose deviations, and contextual changes \cite{pumacay2024colosseumbenchmarkevaluatinggeneralization}. Reinforcement learning, though robust, has a trial-and-error nature and low learning efficiency, making it less applicable in real-world environments.
    
    \item \textbf{Declarative Skill Dilemma.} Recent endeavors to integrate vision-language models (VLM) into robotic systems \cite{liu2024mokaopenworldroboticmanipulation, huang2023voxposercomposable3dvalue,huang2024rekepspatiotemporalreasoningrelational,pan2025omnimanipgeneralroboticmanipulation,cheang2024gr2generativevideolanguageactionmodel} like ReKep \cite{huang2024rekepspatiotemporalreasoningrelational} and OmniManip \cite{pan2025omnimanipgeneralroboticmanipulation} use large multimodal models to generate operational instructions for open-domain tasks. While these models excel in understanding, they lack embodied experience and require constraining outputs to executable actions. This approach forces the language model to handle spatial-temporal reasoning without physical intuition, often resulting in implausible task planning. For instance, in the task “place block A atop block B”, an insufficient spatial understanding (e.g., shape, height) often leads this kind of method to produce fatally flawed action sequences.
\end{itemize}

Unfortunately, in facing these dilemmas, most existing methods \cite{brohan2023rt2visionlanguageactionmodelstransfer,kim2024openvlaopensourcevisionlanguageactionmodel} choose to restrict the capabilities of large multimodal models, forcing them to generate low-level execution commands to drive data-driven downstream control strategies to complete tasks. This enables a certain degree of task execution in open scenarios but essentially represents a compromise in cognitive and executive capabilities. Conversely, suppose we could leverage the strengths of large language/multimodal models in declarative tasks and the advantages of reinforcement learning in procedural tasks without mutual constraints. In that case, robots might achieve significant breakthroughs in handling open-ended tasks. Therefore, the problem we would like to answer is: how to construct a novel framework that allows large multimodal models and reinforcement learning to utilize their respective strengths while supporting each other fully, ultimately maximizing the capability in open-ended scenarios.

In the theory of Central Pattern Generators (CPGs) in the cognitive domain \cite{Grillner1985NeurobiologicalBO,merel2019hierarchical,yuan2023hierarchical}, scientists have proposed that during the process of human beings transitioning from high-level reasoning to low-level control, the brain sends messages, such as walking, specific neurons will always be activated according to specific abstract concepts, regardless of the specific manifestations of the environment, such as appearance, color, size, etc. The activation of these specific neurons regulates the activity level of the motor regions \cite{gazzaniga2006cognitive}. Eventually, precise action behaviors emerge according to the manifestations of the environment, like determining how large a step to take with the left foot or the right foot.
This theory provides an inspiring perspective, suggesting that the communication between high-level reasoning and low-level control might go through a generalized interaction representation with physical intuition and environmental invariance, serving as a symbolic bridge between abstract cognition and embodied execution, rather than taking the form of direct communication.

In this paper, we introduce \textit{RoBridge}—a hierarchical intelligent architecture aimed at general robotic manipulation. As shown in Figure \ref{fig:comparison}, our method maintains the declarative skill of VLM, and releases the procedural skill of reinforcement learning. This architecture innovatively constructs an intelligent system comprising three core elements: 
(1) a high-level cognitive planner (HCP) based on a large-scale pre-trained vision-language model;
(2) an invariant operable representation (IOR) with physical intuition and environmental invariance, serving as a symbolic bridge between abstract cognition and embodied execution; 
(3) a guided embodied agent (GEA) that achieves precise action execution through interaction with the physical world.
By generating IOR through the causal reasoning capabilities of the HCP 
and transforming abstract symbols into operations in physical space via the GEA, RoBridge effectively resolves the challenge of the disconnection between cognition and execution.
While prior works have partially touched on individual ingredients, RoBridge is, to our knowledge, the first holistic system that synergistically unites the declarative power of large vision–language models with the procedural proficiency of reinforcement learning via the invariant operable representation (IOR) without mutual restriction.
Our main contributions are as follows:

\begin{itemize}
    \item We propose the first hierarchical intelligent architecture, RoBridge, for general robotic manipulation, breaking the paradigm dilemma of disconnection between cognitive abstraction and physical execution in traditional methods through a three-tier architecture of brain: HCP, symbolic bridge: IOR, and embodied agent: GEA.
    \item Guided Embodied Agent: We design a Guided Embodied Agent (GEA) capable of converting invariant operable representation (IOR) into concrete execution actions, maintaining superior performance under various interference conditions.
    \item We demonstrate the effectiveness of RoBridge through a series of experiments, showing its ability to generalize to unseen environments. We also find that RoBridge excels at handling new tasks and achieves a 75\% success rate on five new tasks. And it excels in sim-to-real generalization, achieving an average success rate of 83\% with only five real-world data samples per task.
\end{itemize}
\section{Related Work}
\label{sec:related_work}

\textbf{Robotic Manipulation Learning.}
Robotic manipulation has garnered extensive research attention in recent years. A straightforward approach utilizes imitation learning \cite{zhang2018deepimitationlearningcomplex, pomerleau1988alvinn}, a form of supervised learning that maps observations to actions. Early methods achieved high performance by designing effective network architectures, constructing diverse training objectives, and utilizing suitable representations \cite{brohan2023rt1roboticstransformerrealworld, octo_2023,fu2024mobilealohalearningbimanual, wu2023unleashinglargescalevideogenerative}. To enable the deployment of strategies in diverse real-world scenarios, several large-scale robotic datasets have been introduced \cite{embodimentcollaboration2024openxembodimentroboticlearning, khazatsky2024droidlargescaleinthewildrobot, fang2023rh20tcomprehensiveroboticdataset, wang2024all}; however, the scale of these datasets pales in comparison to the expansive variability of real-world environments. Inspired by the success of large pre-trained models \cite{openai2024gpt4technicalreport, touvron2023llamaopenefficientfoundation, alayrac2022flamingovisuallanguagemodel}, researchers have begun fine-tuning vision-language models directly on robotic data to enhance the generalization capabilities of robotic models \cite{brohan2023rt2visionlanguageactionmodelstransfer, kim2024openvlaopensourcevisionlanguageactionmodel, li2024generalistrobotpoliciesmatters,black2024pi0visionlanguageactionflowmodel}. Nonetheless, due to the domain gap between robotic and pre-training data, fine-tuned models often suffer from catastrophic forgetting, resulting in a significant decline in cognitive performance. An alternative approach involves reinforcement learning for robotic manipulation \cite{openai2019solvingrubikscuberobot, yarats2021masteringvisualcontinuouscontrol, wu2022daydreamerworldmodelsphysical, dalal2024planseqlearnlanguagemodelguided}, which can develop robust policies. However, these methods frequently struggle with complex tasks requiring language comprehension. In contrast, our approach transforms observations from the physical world into invariant, physically intuitive representations, integrating reinforcement learning and imitation learning to construct efficient and generalizable embodied agents capable of executing tasks across diverse environments.

\noindent\textbf{Representations for Robotic Manipulation.}
Structured representations are designed to address the generalization challenges in robotic manipulation. In robotic manipulation, structured representation is a method of encoding complex environmental or object information into abstract forms with explicit geometric, semantic, or functional structures to support efficient reasoning, planning, and task execution. The core idea is to capture key information through simplified symbolic elements (such as keypoints \cite{liu2024mokaopenworldroboticmanipulation, pan2025omnimanipgeneralroboticmanipulation,huang2024rekepspatiotemporalreasoningrelational}, 6D poses \cite{wang2023adaaffordlearningadaptmanipulation,ju2024roboabcaffordancegeneralizationcategories,kuang2024ramretrievalbasedaffordancetransfer}, constraints \cite{zhu2023learninggeneralizablemanipulationpolicies,yuan2022sornetspatialobjectcentricrepresentations,cheng2024nodtampgeneralizablelonghorizonplanning,hsu2023whatsleftconceptgrounding}) to reduce problem complexity and enhance generalization capabilities. 
Some methods \cite{black2023zeroshotroboticmanipulationpretrained,du2023learninguniversalpoliciestextguided,zhang2024pivot} predict key frames of the motion process, utilizing them as primitive prompts for low-level control. 
Other approaches \cite{ju2024roboabcaffordancegeneralizationcategories, kuang2024ramretrievalbasedaffordancetransfer} extract interaction trajectories from human videos to tap into diverse data sources, thereby enhancing the generalization to a wide array of previously unseen objects. This, however, often necessitates task-specific annotations. Additionally, some studies \cite{pan2025omnimanipgeneralroboticmanipulation, huang2024rekepspatiotemporalreasoningrelational} leverage the capabilities of vision-language models \cite{openai2024gpt4technicalreport} and foundational models \cite{kirillov2023segment, oquab2024dinov2learningrobustvisual, yang2023setofmarkpromptingunleashesextraordinary} to derive keypoints and 6D poses as representations, integrating these with motion planning for low-level control. Nonetheless, the absence of physical interaction data during pre-training often results in vision-language models generating plans that do not align with real-world physics, thereby limiting their effectiveness in practical applications.
Our approach proposes a physically intuitive representation that does not involve specific execution details, leaving the execution to be handled by underlying embodied agents.

\noindent\textbf{Generalizable Robot Manipulation.}
To enhance model generalization across different environments, particularly for transferring from simulated to real-world settings, existing methods frequently employ domain adaptation and domain randomization techniques. Domain adaptation aims to create additional synthetic images based on existing ones \cite{zhu2024nerfaugdataaugmentationrobotics, ho2021retinaganobjectawareapproachsimtoreal, rao2020rlcycleganreinforcementlearningaware}. In contrast, domain randomization is more commonly used because it only requires randomizing material textures, object poses, camera parameters, and other factors \cite{garcia2023robustvisualsimtorealtransfer,tobin2017domainrandomizationtransferringdeep,strudel2020learningcombineprimitiveskills}. Domain randomization is typically applied within simulated environments, with the trained policies subsequently transferred to real robots. Given its convenience, our approach adopts domain randomization to improve performance. We implement various domain randomization in the simulation environment, achieving high success rates in sim-to-real transfer with as few as five data samples.
\section{Method}
\begin{figure*}
    \centering
    \vspace{-1em}
    \includegraphics[width=0.95\linewidth]{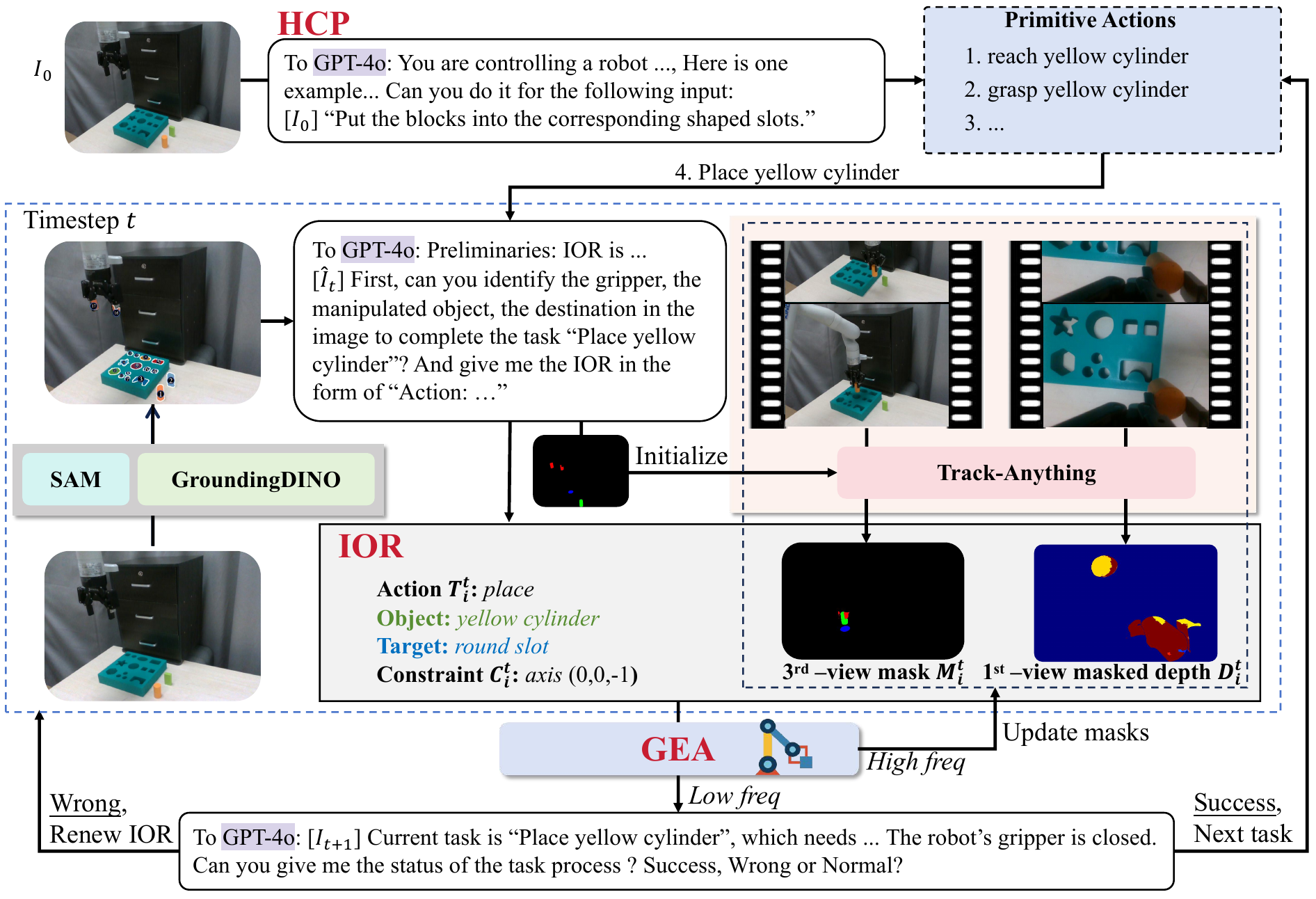}
    \vspace{-0.5em}
    \caption{\textbf{RoBridge overview}. RoBridge adopts a three-layer architecture, consisting of a high-level cognitive planner (HCP), an invariant operable representation (IOR), and a guided embodied agent (GEA). For example, for the instruction ``Put the blocks into the corresponding shaped slots", HCP will first plan and split the task into multiple primitive actions. Then, combined with the APIs composed of the foundation model, it will give IOR, which mainly includes the masked depth of the first perspective, the mask of the third perspective, the type of action, and the constraints. IOR is updated by HCP at a low frequency and track-anything updates the mask at a high frequency. IOR is used as the input of GEA, and GEA performs specific actions until the task is completed.
    }
    \label{fig:framework}
    \vspace{-0.5em}
\end{figure*}

\subsection{Framework}

The framework of RoBridge is illustrated in Figure \ref{fig:framework}. 
It mainly contains three core components: a high-level cognitive planner, an invariant operation representation and a guided embodied agent.
First, the high-level cognitive planner decomposes tasks into primitive actions based on observation and instruction. Then, for each action, the HCP integrates foundation models to generate an invariant operation representation. Finally, the guided embodied agent performs operations utilizing this representation, with the execution process being governed by closed-loop control. 
Detailed descriptions follow.

\noindent\textbf{High-level Cognitive Planner (HCP).}  RoBridge's HCP consists of VLM (such as GPT-4o \cite{openai2024gpt4technicalreport}) and some APIs (such as GroundingDINO \cite{liu2024groundingdinomarryingdino}, SAM \cite{kirillov2023segment} and Track-Anything \cite{yang2023trackanythingsegmentmeets} It is mainly responsible for high-level cognitive planning of tasks. Specifically, given the currently observed RGB Image $I^{h\times w\times 3}$ and instruction $\ell$, we query VLM to decompose the task into several primitive actions, each primitive action $\mathcal{A}_i = \{T_i,\: obj_i,\: des_i\}$, with $T_i$ denoting the type of action, $obj_i$ indicating the name of the manipulated object and $des_i$ indicating the destination or may not exist. The detailed definitions of primitive actions are shown in the Appendix. An example as shown in Figure \ref{fig:framework}, the task is decomposed into: \\
$\{ \mathcal{A}_1 = \{reach,\, yellow\, cylinder,\, none\}, \\
\mathcal{A}_2 =\{grasp,\, yellow\, cylinder,\, none\},\\ 
\mathcal{A}_3 =\{reach,\, round\, slot,\, none\}, \\
\mathcal{A}_4 = \{place,\, yellow\, cylinder ,\, round\, slot\}\}$.\\
This helps the model complete high-level cognitive planning of complex tasks.

\noindent\textbf{Invariant Operable Representation (IOR).} 
For each primitive action $A_i$ and sensors data, we aim to transform them into an invariant operable representation $\mathcal{R}_i$,  which consists of action type $T_i$, the $3^{rd}$-view mask $M_i$, the $1^{st}$-view masked depth $D_i$ and constraints $C_i$.
$M_i$ consists of the $3^{rd}$-view mask of the gripper $M^g_i$, the $3^{rd}$-view mask of the manipulated object $M^o_i$, the $3^{rd}$-view mask of destination $M^d_i$ (if exists). Similarly, $D_i$ consists of the $1^{st}$-view masked depth of the gripper $D^g_i$, the $1^{st}$-view masked depth of the manipulated object $D^o_i$, the $1^{st}$-view masked depth of destination $D^d_i$ (if exists). The constraints $C_i$ include end-effector pose and direction of movement.
The comprehensive representation for each action $A_i$ is given by:
\begin{equation}
\mathcal{R}_i = \{ T_i, M_i, D_i, C_i \}.
\end{equation}
To obtain $\mathcal{R}_i$, 
we first use the APIs GroundingDINO \cite{liu2024groundingdinomarryingdino} and SAM \cite{kirillov2023segment} in HCP to segment objects related to the primitive action.
Subsequently, HCP's VLM determines the final selection of these objects like \cite{yang2023setofmarkpromptingunleashesextraordinary}. Simultaneously, for tasks with directional constraints, such as opening a drawer or turning a faucet, HCP provides corresponding direction $\mathbf{d} \in \mathbb{R}^3$, which is normalized and incorporated into the constraints $C_i$. Integrating this information with sensor data, we ultimately derive $\mathcal{R}_i$.
IOR aims to help RoBridge gain better domain invariance and reduce the impact of environmental and task changes on the model.

\noindent\textbf{Guided Embodied Agent (GEA).}
At each time step, a new $\mathcal{R}^t_i$ is generated. The specific update process will be described in the next closed-loop control section.
For each $\mathcal{R}^t_i$, we need to map it into robotic movement $\mathbf{a}_t$ to proceed with the primitive action. In other words, we need to learn a policy $ \pi \left(\mathcal{R}^t_i\right) \mapsto \mathbf{a}_t $. 
Following PSL \cite{dalal2024planseqlearnlanguagemodelguided}, the primitive action ``reach" typically involves moving the robot's end-effector to a well-defined target position. Unlike tasks such as grasping or placing, which involve complex object interactions and decision-making, ``reach" can be efficiently addressed through motion planning.
For other primitive actions, we integrate reinforcement learning and imitation learning to train a guided embodied agent capable of performing actions effectively, while ensuring robustness and consistent performance under varying input conditions. Detailed descriptions are shown in Section \ref{agent}.

\noindent\textbf{Closed-Loop Control.} In order to maintain the accuracy of information and iteration of primitive actions in dynamic environments, we added closed-loop control. 
Since the speed of each part of the closed-loop control is different, we divide it into high-frequency control and low-frequency control.
For high-frequency control, the update of $\mathcal{R}^t_i$ is performed as follows: first, new sensor data is acquired, and subsequently, Track-Anything \cite{yang2023trackanythingsegmentmeets} is employed to update $M_i^t,D_i^t$, thereby yielding the updated $\mathcal{R}^t_i$.
For low-frequency control, we try to check and get the following three statuses: success, wrong and normal. We use GPT-4o combined with the gripper status to determine whether the task is successful, like \cite{zhang2024pivotrprimitivedrivenwaypointawareworld}. 
We input the following information into GPT-4o for processing: an RGB image $\hat{I}_t$ annotated with task-related object tags, the status of the gripper (indicating whether it is open or closed), and the primitive action. Based on this input, GPT-4o generates a judgment regarding the success or failure of the current action.
If the judgment indicates success, the system proceeds to the next action in the sequence or terminates the task, depending on whether all required actions have been completed. Conversely, if the judgment indicates failure, the system regenerates the input $\mathcal{R}_i$.

\subsection{Guided Embodied Agent (GEA) Training}
\label{agent}
\begin{figure*}
    \centering
    \vspace{-1em}
    \includegraphics[width=1\linewidth]{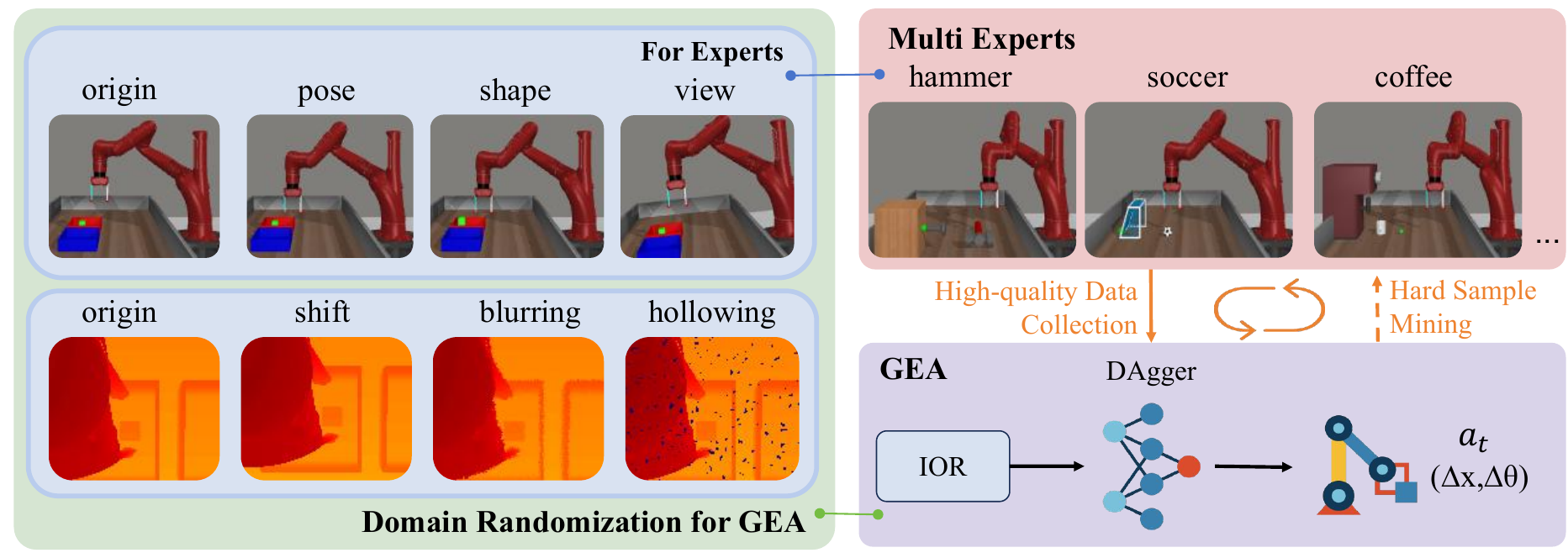}
    \vspace{-1em}
    \caption{\textbf{Guided Embodied Agent(GEA) Training}. The left figure illustrates the domain randomization methods, with the first column showing the original images. The first row employs changes in robotic arm pose, object shape variations, and camera offsets, while the second row uses pixel offsets, depth distortion, hollowing, etc. The domain randomization in the first row is used during expert training, whereas all domain randomization methods are applied during GEA training. In the top right corner, it represents that we train an expert for each task, and the trained experts provide training data for GEA. The bottom right corner shows the training process of GEA, where it takes IOR as input and is trained using DAgger. Data from failed executions is corrected by the experts.
    }
    \label{fig:GEA}
\end{figure*}

We aim to train an agent that achieves a high success rate and strong robustness, ensuring reliable and consistent performance across diverse scenarios, robot configurations, and tasks.
To do this, we used multi-stage training, as shown in Figure \ref{fig:GEA}.

\noindent\textbf{RL Training.} Initially, it is imperative to attain high performance. To this end, we employ reinforcement learning methodologies to train an expert agent $\pi_e$ for each specific task, ensuring that each task is executed proficiently. 
Furthermore, to enhance the robustness of the expert agent, we introduce domain randomization variations in object shape, robotic arm placement, and camera orientation during the training process.

\noindent\textbf{IL Training.} Next, we start training a guided embodied agent $\pi_g$. We use expert agents to generate high-quality data for each task and extract generalized interaction representation $\mathcal{R}$ from the data as the input of $\pi_g$. 
In addition to the domain randomization employed during experts agent training, we further incorporated several domain randomization techniques, including depth distortion, dilation, random shifting, and the modification of masked information through addition and deletion. Such strategies are intended to improve the generalization capabilities and robustness of the trained agent across varying environmental conditions. 
To simulate the noisy depth sensors in the real world, we employ a series of augmentation techniques \cite{dalal2024localpoliciesenablezeroshot} when processing depth images. Specifically, we utilize depth warping, which applies Gaussian shifts to simulate variations in perspective and sensor noise. Additionally, we incorporate Gaussian blurring to mimic the effects of sensor blur and focus issues, which are common in real-world scenarios. Furthermore, we introduce random masking to create artificial holes in the depth maps, simulating occlusions and missing data that frequently occur in practical applications. 
Due to the potential inaccuracies and incomplete coverage inherent in the model-generated masks, we employ techniques such as random offsets and random cropping. These methods are applied to mitigate the risk of the agent becoming overly reliant on the masks.

\noindent\textbf{Continual Skill
Aggregation} 
The agent often experiences compounding errors in imitation learning \cite{zhao2023learningfinegrainedbimanualmanipulation}. To address this, we introduce an iterative optimization strategy using DAgger \cite{ross2011reductionimitationlearningstructured}. Recognizing that online DAgger \cite{agarwal2022leggedlocomotionchallengingterrains} can cause instability and offline DAgger updates slowly in multitask settings, we developed an adaptive sampling mechanism for offline DAgger. As detailed in Algorithm \ref{Dagger}, this mechanism adjusts the sampling frequency based on task complexity: more samples for challenging ones. Initially, tasks are weighted equally. A function $f$ maps rewards to values to adjust task weights. In each iteration, policy $\pi_g$ is trained on current datasets, tasks are sampled by weight, and rewards are transformed by $f$ to update weights. Failures are recorded, and experts $\pi_e$ provide corrective data for the task dataset.

\begin{algorithm}
\caption{Adaptive Sampling DAgger}
\label{Dagger}
\begin{algorithmic}[1]
\STATE Initialize equal weights $w_i$ for all tasks $\{T_i\}$
\STATE Define a  piecewise function $f: \text{reward} \rightarrow \text{value}$
\STATE Collect initial dataset $D_i$ for each task $T_i$ with equal size

\WHILE{not converged}
    \STATE Train policy $\pi_g$ using datasets $\{D_i\}$
    \STATE Sample $n$ tasks according to weights $w_i$
    \STATE Test policy $\pi_g$ with $n$ tasks
    
    \FOR{each task $T_i$}
        \FOR{each reward in task $T_i$ tests}
            \STATE $w_i' += f(\text{reward})$
        \ENDFOR
        \STATE $w_i' = w_i' / \text{number of tests for } T_i$
        \STATE Record all failure data, let expert $\pi_e$ generate correct data $D'_i$
        \STATE $D_i = D_i + D'_i$
    \ENDFOR
    
    \STATE Update $w_i = w_i'$
\ENDWHILE

\end{algorithmic}
\end{algorithm}
\section{Experiments}
\label{sec:exp}

We first introduce the experimental setup \ref{exp_setting} along with the benchmarks and baselines \ref{baselines}. We present the main results \ref{results} and demonstrate performance on unseen tasks \ref{unseen_tasks}. Additionally, we conduct ablation studies \ref{ablation}.

\begin{table*}[t]
    \centering
    \small
    \caption{Metaworld \cite{yu2021metaworldbenchmarkevaluationmultitask} Benchmark Results (\%). }
    \vspace{-0.8em}
    \begin{tabular}{l|ccccc|c}
    \toprule
        Model &  MT50 &  Unseen Background & Unseen Light & Unseen Color & Unseen Camera Pose& Mean\\ \midrule 
        DRQ-v2 \cite{yarats2021masteringvisualcontinuouscontrol}& 24.0 & 24.6 & 17.4 & 22.2 & 19.8 & 21.60 \\
        SayCan \cite{ahn2022icanisay}& 77.8 & 26.6 & 27.8 & 12.8 & 10.4 & 31.08\\
        PSL \cite{dalal2024planseqlearnlanguagemodelguided}& 72.4 & 28.8 & 37.8 & 42.8 & 14.4 & 45.24\\ 
        ManipGen \cite{dalal2024localpoliciesenablezeroshot}& 73.2 & 75.4 & 70.4& 72.4 & 62.8& 70.84 \\ 
        ReKep \cite{huang2024rekepspatiotemporalreasoningrelational}& 31.4 &32.8 & 31.8& 32.2& 32.6& 32.16\\ \midrule 
        RoBridge & $\mathbf{85.4}$  & $\mathbf{84.6}$ & $\mathbf{82.6}$ & $\mathbf{83.2}$ & $\mathbf{74.8}$ & $\mathbf{82.12}$ \\
        
    \bottomrule
    \end{tabular}
    
    \vspace{-0.5em}
    \label{tab:metaworld}
\end{table*}

\begin{table*}[t]
    \centering
    \small
    \caption{Real-World Experimental Results (\%). }
    \vspace{-0.8em}
    \begin{tabular}{l|cccccc|c}
    \toprule
        Model &  \multicolumn{2}{c}{Pick Up} & \multicolumn{2}{c}{Sweep} & Press Button  & Open Drawer & Mean\\ 
        & Seen & Unseen & Seen & Unseen & & & \\
         \midrule 
        ManipGen\cite{dalal2024localpoliciesenablezeroshot}& 10 & 0 & 0 & 0& 30 & 0 & 6.7\\
        RDT \cite{liu2024rdt1bdiffusionfoundationmodel}& 20 & 0 & 0 & 0 & 25 & 0 & 7.5\\ 
        $\pi_0$ \cite{black2024pi0visionlanguageactionflowmodel}& 40  & 10 & 30 &0 & 10 & 20 & 18.3\\ 
        $\pi_0$-fast \cite{black2024pi0visionlanguageactionflowmodel}& 35 & 10 & 25 & 0 & 30 & 10 & 18.3 \\ 
        RAM \cite{kuang2024ramretrievalbasedaffordancetransfer} & 60 & 55 & 0 & 0 & 80 & 70 & 44.17\\  
        ReKep \cite{huang2024rekepspatiotemporalreasoningrelational}& 80 & $\mathbf{85}$ & 40 & 30 & 5 & 55 & 49.2\\ \midrule 
        RoBridge &  $\mathbf{85}$ & $\mathbf{85}$ & $\mathbf{70}$ &  $\mathbf{65}$ & $\mathbf{95}$  & $\mathbf{100}$ & $\mathbf{83.3}$ \\
    \bottomrule
    \end{tabular}
    
    \vspace{-0.5em}
    \label{tab:real_world}
\end{table*}

\subsection{Experiment Settings}
\label{exp_setting}
\textbf{Architecture and Training.} For each task, we employ a separate reinforcement learning (RL) policy, denoted as expert $\pi_e$. We train $\pi_e$ using DRQ-v2 \cite{yarats2021masteringvisualcontinuouscontrol}, a state-of-the-art RL algorithm, which takes as input both RGB images and the robot's state, alongside a task-specific one-hot vector representation, and outputs low-level actions.  $\pi_g$ employs a network architecture identical to that of DRQ-v2, with the exception of the absence of the critic network. This framework utilizes the Invariant Operable Representation (IOR) as input, where primitive actions are represented by one-hot vectors, and similarly outputs low-level actions.

\noindent\textbf{Hardware Setup.} In conducting real-world experiments, we utilize a Kinova Gen3 robotic arm. We also employ two Realsense D435i cameras: one mounted on the wrist to provide a first-person perspective, and the other one offers a third-person view.

\subsection{Benchmark and Baselines} 
\label{baselines}
\noindent\textbf{Simulation Benchmark.} We conducted experiments in two simulation environments: Metaworld \cite{yu2021metaworldbenchmarkevaluationmultitask} and Robosuite \cite{zhu2025robosuitemodularsimulationframework}. Metaworld provides a diverse set of tasks, and we utilized it for both training and testing across 50 available tasks. In addition, we carefully selected 35 tasks for training and five distinct tasks for zero-shot testing, ensuring that there was no correlation between the training and testing task sets. The experiments on Robosuite are shown in the Appendix.

\noindent\textbf{Real-world Evaluation.} We conducted real-world experiments involving the following tasks: (1) ``Pick up": accurately grasp and lift the specified object from the table. (2) ``Sweep": move the object to a designated location. (3) ``Press button": depress the button fully to achieve closure. (4) ``Open Drawer": extend the drawer by a minimum distance of 10 cm. ``Pick up" and ``Sweep" tasks are tested with unseen objects to further evaluate generalization capability.
We additionally designed a multi-stage task, which needs to pick up blocks and insert them into a correspondingly shaped groove. The task is designed to evaluate the model's capability and stability in handling long-horizon sequences. We set up four stages: ``pick cylinder", ``place cylinder", ``pick cuboid", and ``place cuboid", and measured the length of the execution process.

\noindent\textbf{Baselines.} 
DrQ-v2 \cite{yarats2021masteringvisualcontinuouscontrol} is a state-of-the-art reinforcement learning method. We adapted a multi-task DrQ-v2 using RGB inputs, robot proprioception, and one-hot encoded tasks. SayCan \cite{ahn2022icanisay} uses a large language model (LLM) for skill planning. We used DrQ-v2 models trained on separate tasks as SayCan's skill library. PSL \cite{dalal2024planseqlearnlanguagemodelguided} organizes skills as actions, similar to SayCan. ManipGen \cite{dalal2024localpoliciesenablezeroshot} extends PSL with DAgger and domain randomization. We expanded PSL and ManipGen's skill libraries for our tasks. 
RAM \cite{kuang2024ramretrievalbasedaffordancetransfer} is a retrieval-based, zero-shot robotic manipulation framework that transfers 2D affordances from diverse out-of-domain data to 3D executable actions.
ReKep \cite{huang2024rekepspatiotemporalreasoningrelational} employs keypoint representations for task planning. Due to its initial suboptimal performance, we enhanced prompts with additional task specifications and constraints. We also compared popular end-to-end methods, including RDT \cite{liu2024rdt1bdiffusionfoundationmodel}, $\pi_0$ \cite{black2024pi0visionlanguageactionflowmodel}, and its autoregressive version, $\pi_0$-fast. For real-world testing of RDT, $\pi_0$, and our RoBridge, we collected 5 demonstrations per task for fine-tuning.

\subsection{Results Analysis}
\label{results}

\textbf{Simulation Results.} We conducted tests on 50 tasks from Metaworld, and changed the background, lighting, object colors, and camera poses to evaluate generalization capabilities. The average success rate is presented in the last column. As shown in Table~\ref{tab:metaworld}, our method achieved the best performance across various tests, with an average success rate of 82.12\%, which is 11.28\% higher than the best baseline. This demonstrates the effectiveness and robustness of our approach.

\begin{table}[!t]
    \centering
    \small
    \vspace{-0.5em}
    \caption{Real-World Long-horizon Experimental Results (\%). }
    \vspace{-0.8em}
    \begin{tabular}{l|cccccc|c}
    \toprule
        Model &  1 & 2 & 3  & 4 & Avg. Len.\\ 
         \midrule 
        ManipGen\cite{dalal2024localpoliciesenablezeroshot}& 20 & 0& 0& 0& 0.2 \\
        RDT \cite{liu2024rdt1bdiffusionfoundationmodel}& 30 & 0 & 0 & 0&  0.3 \\ 
        $\pi_0$ \cite{black2024pi0visionlanguageactionflowmodel}& 45 & 5 & 0& 0& 0.5 \\ 
        $\pi_0$-fast  \cite{black2024pi0visionlanguageactionflowmodel}& 30& 10& 0& 0& 0.4 \\ 
        ReKep \cite{huang2024rekepspatiotemporalreasoningrelational}& $\mathbf{100}$ & 40& 25& 5& 1.7 \\ \midrule 
        RoBridge &  $\mathbf{100}$ & $\mathbf{80}$ &  $\mathbf{70}$ & $\mathbf{50}$ & $\mathbf{3.0}$    \\
    \bottomrule
    \end{tabular}
    
    \vspace{-1em}
    \label{tab:long_horizon}
\end{table}

\begin{table*}[!t]
    \centering
    \small
    \caption{Experimental Results (\%) on Unseen Tasks. `-' indicates that the method cannot handle this task due to its underlying principles.}
    \vspace{-0.8em}
    \begin{tabular}{l|ccccc|c}
    \toprule
        Model &  Bin Picking & Pick out & Handle press & Plate Slide & Sweep Into & Mean\\ \midrule 
        DRQ-v2 \cite{yarats2021masteringvisualcontinuouscontrol}& - & - & - & - & - & -\\
        SayCan \cite{ahn2022icanisay}& - &  -& - & - & - & - \\
        PSL \cite{dalal2024planseqlearnlanguagemodelguided}& 0 & 0 & 60 & 0 & 0 & 12 \\ 
        ManipGen \cite{dalal2024localpoliciesenablezeroshot}& 20 & 0 & 60 & 10 & 0 & 18 \\ 
        ReKep \cite{huang2024rekepspatiotemporalreasoningrelational}& 0 & $\mathbf{60}$ & 50 & 0 & 30& 28\\ \midrule 
        RoBridge &  $\mathbf{70}$ & 45 & $\mathbf{100}$ & $\mathbf{90}$ & $\mathbf{70}$ & $\mathbf{75}$\\
        
    \bottomrule
    \end{tabular}
    
    \label{tab:unseen_tasks}
\end{table*}

\begin{table*}[!t]
    \centering
    \small
    \caption{Ablations Study Results (\%). }
    \vspace{-0.8em}
    \begin{tabular}{l|ccccc|c}
    \toprule
        Model &  MT50 & \makecell{Unseen\\Background} & \makecell{Unseen\\Light} & \makecell{Unseen\\Color} & \makecell{Unseen\\Camera Pose}& Mean\\ \midrule 
        RoBridge & $\mathbf{85.4}$  & $\mathbf{84.6}$ & $\mathbf{82.6}$ & $\mathbf{83.2}$ & $\mathbf{74.8}$ & $\mathbf{82.12}$ \\ \midrule
        RoBridge w/ original depth & 87.2 & 80.3 & 84.2 & 84.2 & 60.8 & 79.74 \\ 	
        RoBridge w/ DINOv2 & 62.2 & 57.2 & 47.6 & 54.2 & 30.6 & 50.36 \\
        RoBridge w/ keypoints & 80.2 & 62.2 & 55.4 & 58.4 & 43.2 & 59.88 \\      
        RoBridge w/ language-only & 84.8 & 34.4 & 26.8 & 30.2 & 20.2 & 39.28 \\   
        RoBridge w/ smaller VLM & 65.4 & 62.8 & 60.4 & 66.2 & 64.4 & 63.84 \\
        RoBridge w/o DAgger & 68.4 & 69.0 & 68.6 & 69.4 & 50.2 & 65.12 \\
        RoBridge w/o online learning & 75.4 & 75.2 & 76.4 & 74.8 & 66.6 & 73.68 \\
        RoBridge w/o domain randomization & 62.8 & 62.8 & 62.4 & 62.2 & 30.2 & 56.08 \\
         
    \bottomrule
    \end{tabular}
    
    \label{tab:ablation}
\end{table*}

\begin{figure*}[!t]
    \centering
    \includegraphics[width=1\linewidth]{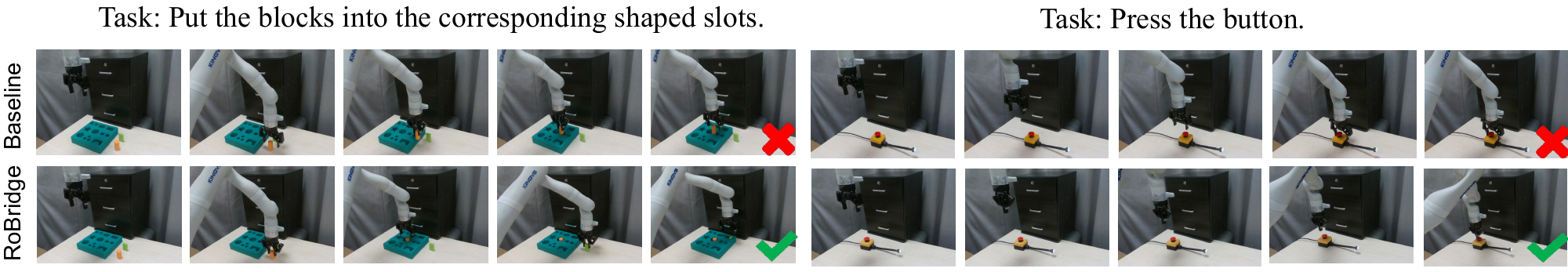}
    \caption{ Demonstrations show the execution process of RoBridge (second row) and baselines $\pi_0$ (left in the first row), Rekep (right in the first row). 
    }
    \vspace{-1.5em}
    \label{fig:visualization}
\end{figure*}

\noindent\textbf{Real-World Results.} The real-world results are shown in Table~\ref{tab:real_world} and Table~\ref{tab:long_horizon}. Our method achieved the best results, with an average success rate of 83.3\% across four tasks and an average length of 3.0 in the long-horizon task. RoBridge's performance on long-horizon tasks demonstrates its combined planning and execution capabilities.
We also show qualitative results, which are shown in Figure \ref{fig:visualization}. The first row displays the tasks process completed by Rekep and $\pi_0$, respectively. The second row illustrates the process completed by RoBridge.
In end-to-end models, $\pi_0$ achieved relatively good success rates. However, it exhibited instability when handling long-horizon tasks and is easy to failure.
ReKep struggles with physics knowledge, such as not knowing that the gripper needed to be closed to press the button. 
Our model effectively solves the problems of these baselines. More visualizations are shown in the Appendix.

\vspace{-1em}
\subsection{Zero-shot Unseen Tasks Generalization}
\label{unseen_tasks}
We conducted tests on five novel tasks that are unrelated to those used during training. As shown in Table~\ref{tab:unseen_tasks}, RoBridge achieved an average success rate of 75\% across these new tasks. This demonstrates that our method can effectively execute tasks beyond those seen during training, eliminating the need to collect data for each specific task, thereby reducing data collection costs.

\subsection{Ablation Study}
\label{ablation}
We conducted ablation experiments on the components of the IOR and the training of the GEA to understand their significance.
\cite{dalal2024localpoliciesenablezeroshot}  used the original depth instead of the 1st masked depth, and \cite{zhou2025dinowmworldmodelspretrained} used features obtained from DINOv2 \cite{oquab2024dinov2learningrobustvisual} as inputs instead of the mask in 3rd view. We conducted ablations on our modules using these two methods. As shown in the second and third rows of Table \ref{tab:ablation}, the original depth might be easier for the model to learn due to the presence of more details, resulting in a higher success rate on the MT50 tasks. However, the success rate drops sharply when there are changes in camera pose. The performance of DINOv2 was not satisfactory, which we think is due to its pre-trained features not being fully suitable for robotics. The sharp decline in performance for both methods under unseen camera poses indicates that these representations lack good invariance, making it difficult to achieve generalization.
The comparison with using keypoints and language shows that RoBridge's success is not only due to richer supervision but also because the IOR design offers better invariance and intuitiveness. The degradation caused by using a smaller VLM indicates that the capability of VLM is necessary to effectively understand instructions and visual inputs.
We also conducted ablation studies on the methods employed in GEA training, as shown in the last three rows of Table \ref{tab:ablation}. Each method was found to be critically important in enhancing the training effectiveness of the GEA.

\section{Conclusion}
\label{sec:conclusion}

In this paper, we introduce RoBridge, a novel hierarchical intelligent architecture designed to enhance robotic manipulation by bridging the gap between high-level cognitive planning and low-level physical execution. The architecture integrates a high-level cognitive planner, an invariant operable representation, and a guided embodied agent, demonstrating significant advancements in task generalization and execution robustness. Through extensive experiments, RoBridge has shown superior performance and strong zero-shot generalization capabilities in unknown environments and novel tasks.

\section*{Acknowledgements}
This work is supported by Scientific Research Innovation Capability Support Project for Young Faculty under Grant No.ZYGXQNJSKYCXNLZCXM-I28, National Natural Science Foundation of China (NSFC) under Grants No.62476293 and 62272494, Shenzhen Science and Technology Program under Grant No.GJHZ20220913142600001, Nansha Key R\&D Program under Grant No.2022ZD014, General Embodied AI Center of Sun Yat-sen University, and China Postdoctoral Science Foundation under Grant No.2025M771522.

\clearpage
{
    \small
    \bibliographystyle{ieeenat_fullname}
    \bibliography{main}
}

\appendix
\phantomsection
\maketitlesupplementary

\centerline{\textbf{SUMMARY OF THE APPENDIX}}

This appendix contains additional details for this paper. The appendix is organized as follows:

\begin{itemize}
    \item \S\ref{limitations} provides \textbf{Limitations} of our work.
    \item \S\ref{experiment_details} provides \textbf{Experiment Details}.
    
    \item \S\ref{results} shows more \textbf{Visualization}.
   
\end{itemize}

\section{Limitations}
\label{limitations}
RoBridge still presents opportunities for further improvement. As a hierarchical framework, it is susceptible to the performance of any individual module. Enhancements in the visual understanding capabilities of the high-level planning module and improvements in the precision of execution in the low-level control module could significantly boost RoBridge's overall performance. Furthermore, while currently limited to manipulating objects with simple shapes, RoBridge could be extended to handle a wider variety of shapes in the future, including soft or tiny objects.

\section{Experiment Details}
\label{experiment_details}
\subsection{Primitive Actions}
The detailed definitions of primitive actions are shown in Table~\ref{tab:actions}.
\begin{table}[ht]
\centering
\caption{The list of primitive actions and their description.}
\resizebox{1.0\linewidth}{!}{
\begin{tabular}{@{\hspace{0.2ex}}l@{\hspace{0.5ex}}|@{\hspace{0.5ex}}c@{}}
\toprule
\textbf{Action}  & \textbf{Description}\\ \toprule
 grasp & Securely hold an object to control its position. \\ 
place & Put an object at a specific location. \\
press & Apply force to an object to activate or transform it. \\
push & Exert force on an object to move it away from a specific direction. \\ 
pull & Apply force to draw an object closer from a specific direction. \\
open & Adjust an object to allow access or exposure. \\
close & Adjust an object to restrict access or seal it. \\ 
turn & Rotate an object to change its orientation. \\ 
reach & Approach an object or a designated location. \\ 
 \bottomrule

\end{tabular}
}
\label{tab:actions}
\end{table}

\subsection{Training Details}
\label{app:training_details}
\begin{figure*}[!t]
    \centering
    \includegraphics[width=0.8\linewidth]{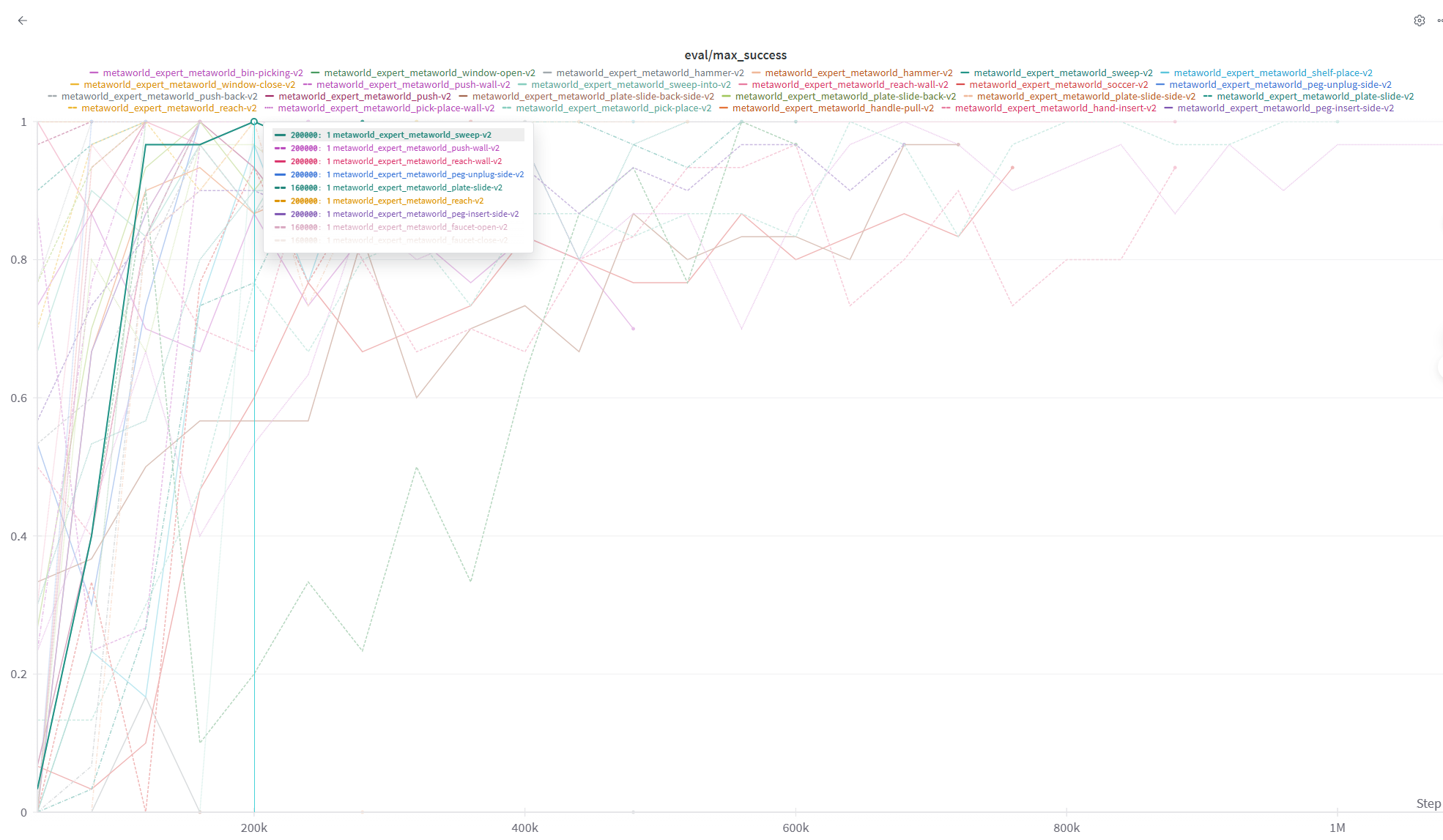}
    \caption{RL training details.
    }
    \label{fig:RL_training}
\end{figure*}
In our experiments on MetaWorld, we conducted training for 1M steps, with failure data sampling occurring every 100k steps. We added the RL training curve. As shown in Figure \ref{fig:RL_training}, most tasks can be trained in 200k timesteps. The success rate of $\pi_e$ in the training scenario is 90.8\%. During real-world experiments, we fine-tuned the model for 2k steps. We will resize the image to 168$\times$168 and feed it to GEA for faster speed.  Training $\pi_e$ and $\pi_g$ on an A100 GPU took 25 and 30 GPU hours, respectively. Finetuning on real data took just 1 GPU hour. Inference needs 6 GB GPU memory (except GPT-4o). Only GEA and Track-Anything run every frame, taking 60-80ms. VLM and SAM+GroundingDINO only run in the first frame of each primitive action, with VLM at 0.3s and SAM+GroundingDINO at 1s per run.

\subsection{Robosuite Benchmark Results}
Robosuite encompasses a suite of contact-rich robotic manipulation tasks, emphasizing high-fidelity rendering and realistic physical control. This environment allows us to evaluate the potential effectiveness of our approach in real-world scenarios. Table~\ref{tab:robosuite} shows the results of Robosuite, RoBridge achieved the highest success rates in all cases, indicating that our method is also effective for tasks involving contact-rich interactions.
\begin{table*}[t]
    \centering
    \small
    \caption{Robosuite Benchmark Results(\%). }
    \begin{tabular}{l|cccccc|c}
    \toprule
        Model &  RS-Bread & RS-Can & RS-Milk & RS-Cereal & RS-NutRound & RS-NutSquare & Mean\\ \midrule 
        DRQ-v2 & 52 & 32 & 2 & 0 & 6 & 2 & 15.7 \\
        RAPS  & 0 & 0 & 0 & 0 & 0 & 0 & 0\\
        TAMP  & 90 & 100 & 85 & 100 & 40 & 35 & 75\\
        SayCan & 93 &  100 & 90 & 63 & 56 & 27 & 71.5 \\
        PSL & 100 & 100 & 100 & 100 & 98 & 97 & 99.2 \\ \midrule 
        RoBridge &  100 & 100 & 100 & 100 & 100 & 100 & 100\\
        
    \bottomrule
    \end{tabular}
    
    \vspace{-1em}
    \label{tab:robosuite}
\end{table*}

\subsection{MetaWorld Benchmark Results}
We show in detail the specific success rate of our method on MT50 (the 50 tasks of the metaworld) in the Table~\ref{mt50}.

\begin{table*}[ht]
\centering
\caption{Detail Results of Metaworld Tasks. Each task is tested 10 times. }
\begin{tabular}{>{\raggedright\arraybackslash}p{4cm} >{\raggedright\arraybackslash}p{1cm} >{\raggedright\arraybackslash}p{4cm} >{\raggedright\arraybackslash}p{1cm} >{\raggedright\arraybackslash}p{4cm} >{\raggedright\arraybackslash}p{1cm}}
\toprule
\textbf{Task} & \textbf{Success} & \textbf{Task} & \textbf{Success} & \textbf{Task} & \textbf{Success} \\
\midrule
assembly             & 10 & button-press-topdown & 10 & door-unlock          & 10 \\
basketball           & 0  & button-press-topdown-wall & 8  & hand-insert          & 10 \\
bin-picking          & 9  & button-press         & 10 & drawer-close         & 10 \\
box-close            & 5  & button-press-wall    & 10 & drawer-open          & 10 \\
coffee-button        & 10 & coffee-pull          & 10 & faucet-open          & 10 \\
coffee-push          & 10 & dial-turn            & 9 & faucet-close         & 10 \\
disassemble          & 6 & door-close           & 10 & hammer               & 2  \\
door-lock            & 8  & door-open            & 10 & handle-press-side    & 10 \\
handle-press         & 10 & handle-pull-side     & 5  & handle-pull          & 10 \\
lever-pull           & 2  & peg-insert-side      & 9  & pick-place-wall      & 10 \\
pick-out-of-hole     & 5 & reach                & 10 & push-back            & 10 \\
push                 & 7 & pick-place           & 10 & plate-slide          & 9 \\
plate-slide-side     & 9 & plate-slide-back     & 8 & plate-slide-back-side & 9 \\
peg-unplug-side      & 9  & soccer               & 7 & stick-push           & 8  \\
stick-pull           & 8  & push-wall            & 10 & reach-wall           & 10 \\
shelf-place          & 9  & sweep-into           & 10 & sweep                & 7 \\
window-open          & 10 & window-close         & 10 & & \\
\midrule
\textbf{Mean Success Rate} & \textbf{85.4} & & & & \\
\bottomrule
\end{tabular}
\label{mt50}
\end{table*}

\section{Visualization}
\label{results}
\subsection{DINOv2 Visualization}
\begin{figure*}[!t]
    \centering
    \includegraphics[width=1\linewidth]{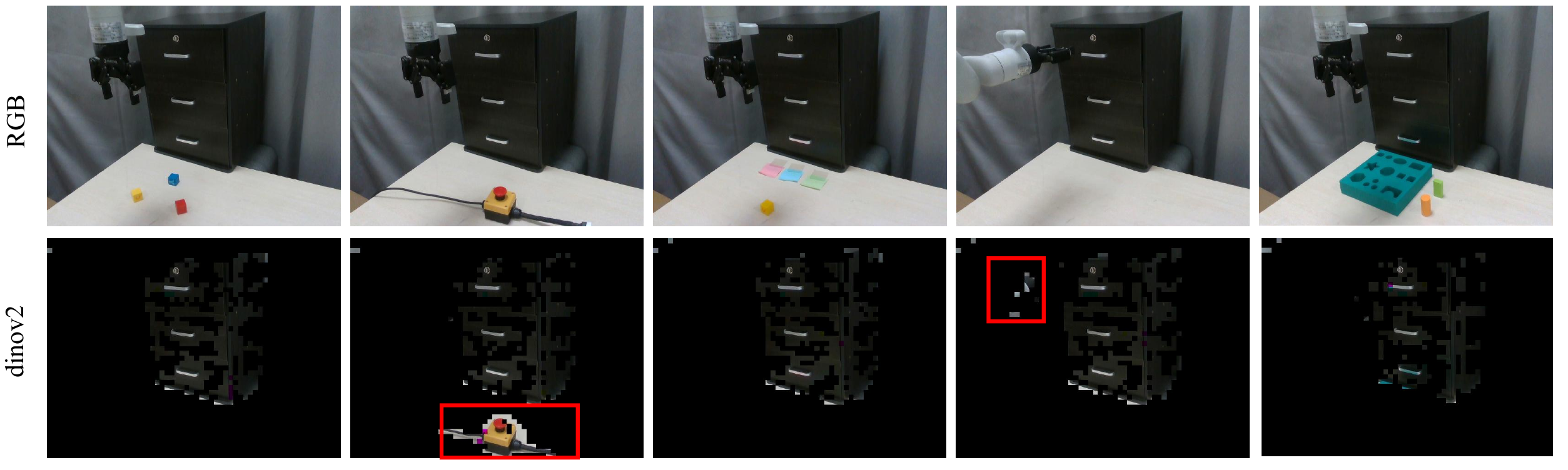}
    \vspace{-2em}
    \caption{ Feature visualization of DINOv2.
    }
    \label{fig:dinov2}
\end{figure*}

We conducted an in-depth analysis of why DINOv2 performs poorly as a feature extractor. By visualizing DINOv2's attention areas, as shown in Figure~\ref{fig:dinov2}, we found that it predominantly focuses on common objects, such as drawers, and does not pay much attention to robotic arms or small objects. It only partially attends to slightly more prominent objects, like buttons, or when a robotic arm is near a drawer. Therefore, it is not well-suited for robotic manipulation tasks.

\subsection{Results}
\begin{figure*}[!t]
    \centering
    \includegraphics[width=1\linewidth]{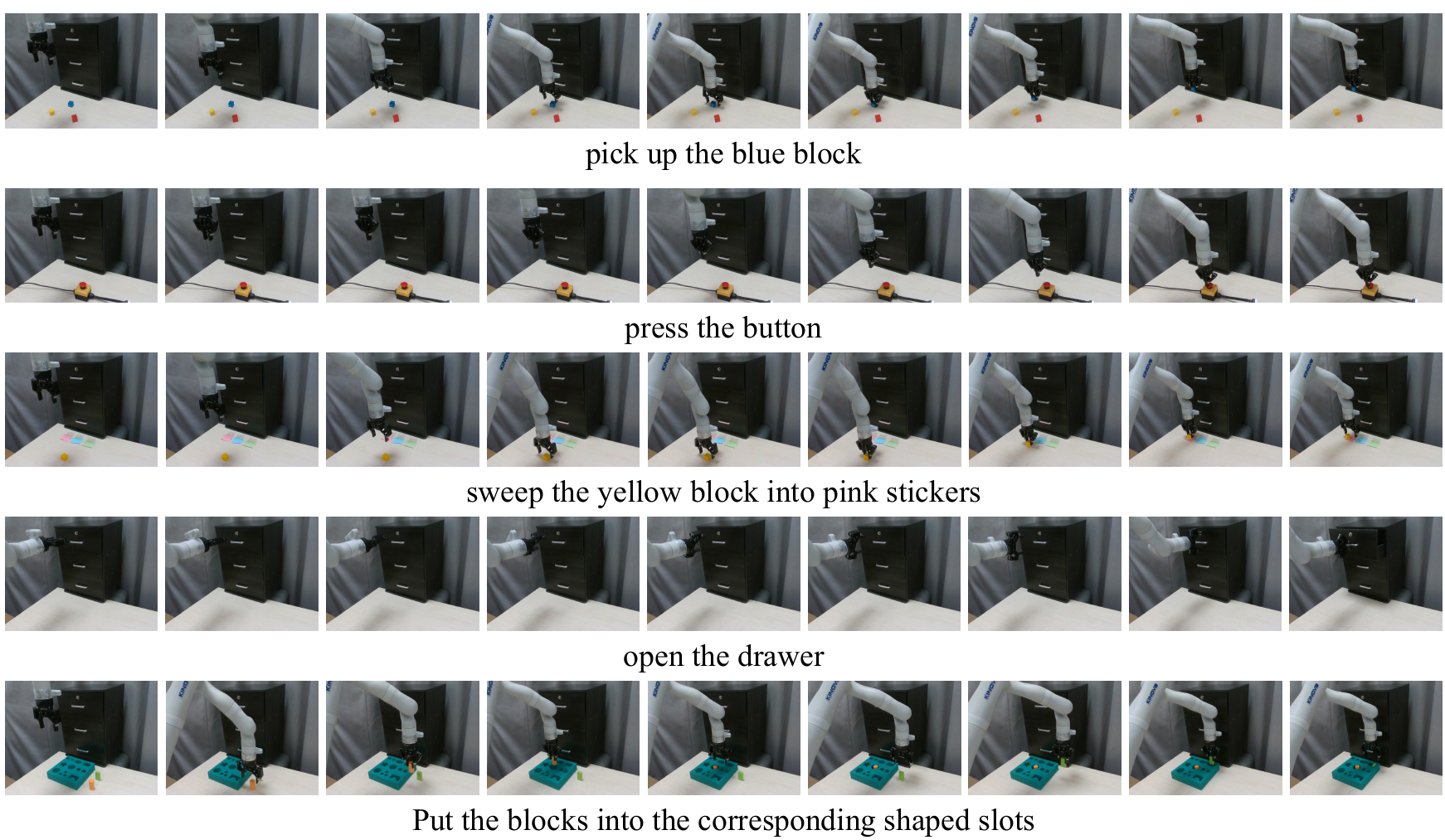}
    \vspace{-2em}
    \caption{ Demonstrations of real-world experiments.
    }
    \label{fig:more_results}
\end{figure*}

We show more demonstrations of real-world experiments in the Figure~\ref{fig:more_results}. 

\subsection{Failure Case Analysis}
\label{failure}
\begin{figure*}[!t]
    \centering
    \vspace{-0.2em}
    \includegraphics[width=1\linewidth]{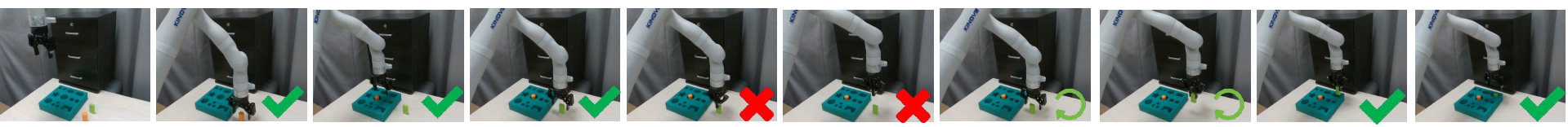}
    \vspace{-1.5em}
    \caption{A demonstration shows how RoBridge recovers from failure. When it fails, it re-plans and executes.
    }
    \label{fig:retry}
    \vspace{-1em}
\end{figure*}
This section investigates the causes of failure in the operation of RoBridge. Our findings indicate that the majority of failures are attributable to the loss of masks due to occlusion or overlap, as well as positional deviations during execution. Fortunately, we observed that RoBridge is capable of correcting errors to a certain extent. As illustrated in Figure~\ref{fig:retry}, when the task failed during the attempt to grasp the second block, RoBridge was able to replan and successfully complete the task on the second attempt.

\end{document}